\documentclass{article}     % uncomment this line when writing your paper

\usepackage{arxiv}
\usepackage[utf8]{inputenc} % allow utf-8 input
\usepackage[T1]{fontenc}    % use 8-bit T1 fonts
\usepackage{hyperref}       % hyperlinks
\usepackage{url}            % simple URL typesetting
\usepackage{booktabs}       % professional-quality tables
\usepackage{amsfonts}       % blackboard math symbols
\usepackage{nicefrac}       % compact symbols for 1/2, etc.
\usepackage{microtype}      % microtypography
\usepackage{lipsum}
\usepackage{graphicx}
\usepackage{caption}
\usepackage{subcaption}
\usepackage{multirow}

\usepackage{graphics} % for pdf, bitmapped graphics files
\usepackage{epsfig} % for postscript graphics files
\usepackage{mathptmx} % assumes new font selection scheme installed
\usepackage{times} % assumes new font selection scheme installed
\usepackage{amsmath} % assumes amsmath package installed
\usepackage{amssymb}  % assumes amsmath package installed
\title{Evolutionary Morphology Towards Overconstrained Locomotion via Large-Scale, Multi-Terrain Deep Reinforcement Learning}
\author{
  Yenan Chen$^{\#,1}$, Chuye Zhang$^{\#,1}$, Pengxi Gu$^{\#,1}$, Jianuo Qiu$^{1}$, Jiayi Yin$^{2}$, Nuofan Qiu$^{2}$, Guojing Huang$^{1}$, \\
  \textbf{Bangchao Huang}$^{1}$, \textbf{Zishang Zhang}$^{1}$, \textbf{Hui Deng}$^{1}$, \textbf{Wei Zhang}$^{3}$\\
  $^{1}$Department of Mechanical and Energy Engineering\\
  $^{2}$School of Design\\
  $^{3}$Department of System Design and Intelligent Manufacturing\\
  Southern University of Science and Technology\\
  Shenzhen, China 518055\\
  \texttt{$^{\#}$Contributed equally as the co-first authors.}\\
  \And
  Fang Wan\thanks{Corresponding Author.}\\
  School of Design\\
  Southern University of Science and Technology\\
  Shenzhen, China 518055\\
  \texttt{wanf@sustech.edu.cn}\\
  \And
  Chaoyang Song$^{*}$\\
  Department of Mechanical and Energy Engineering\\
  Southern University of Science and Technology\\
  Shenzhen, China 518055\\
  \texttt{songcy@ieee.org}\\
}
\begin{document}
\maketitle
%%%%%%%%%%%%%%%%%%%%%%%%%%%%%%%%%%%%%%%%%
\begin{abstract}

    While the animals' Fin-to-Limb evolution has been well-researched in biology, such morphological transformation remains under-adopted in the modern design of advanced robotic limbs. This paper investigates a novel class of overconstrained locomotion from a design and learning perspective inspired by evolutionary morphology, aiming to integrate the concept of `intelligent design under constraints' - hereafter referred to as constraint-driven design intelligence - in developing modern robotic limbs with superior energy efficiency. We propose a 3D-printable design of robotic limbs parametrically reconfigurable as a classical planar 4-bar linkage, an overconstrained Bennett linkage, and a spherical 4-bar linkage. These limbs adopt a co-axial actuation, identical to the modern legged robot platforms, with the added capability of upgrading into a wheel-legged system. Then, we implemented a large-scale, multi-terrain deep reinforcement learning framework to train these reconfigurable limbs for a comparative analysis of overconstrained locomotion in energy efficiency. Results show that the overconstrained limbs exhibit more efficient locomotion than planar limbs during forward and sideways walking over different terrains, including floors, slopes, and stairs, with or without random noises, by saving at least 22\% mechanical energy in completing the traverse task, with the spherical limbs being the least efficient. It also achieves the highest average speed of 0.85m/s on flat terrain, which is 20\% faster than the planar limbs. This study paves the path for an exciting direction for future research in overconstrained robotics leveraging evolutionary morphology and reconfigurable mechanism intelligence when combined with state-of-the-art methods in deep reinforcement learning. (Project GitHub: \href{https://github.com/ancorasir/BennettWheelLegRL}{https://github.com/ancorasir/BennettWheelLegRL})

\end{abstract}
%%%%%%%%%%%%%%%%%%%%%%%%%%%%%%%%%%%%%%%%%
%%%%%%%%%%%%%%%%%%%%%%%%%%%%%%%%%%%%%%%%%%%%%%%%%%%%%%%%%%%%%%%%%%%%%%%%%%%%%%%%
\section{Introduction}
\label{sec:Intro}
%%%%%%%%%%%%%%%%%%%%%%%%%%%%%%%%%%%%%%%%%%%%%%%%%%%%%%%%%%%%%%%%%%%%%%%%%%%%%%%%

    Evidence from paleontology~\cite{Clack2009FintoLimb}, developmental biology~\cite{Clack2009FintoLimb}, and genetic research~\cite{Weaver2010} show a rich body of literature and research on the fin-to-limb evolution that explains how fish evolved into four-limbed vertebrates~\cite{Clack2009FintoLimb, Weaver2010}. While animals evolve in biological morphology in adaptation to environmental changes for survival, engineers are inversely posed with the challenge of searching for an optimal solution. By analogy to the genetic expression of DNA to biological structures, the loop closure formulation holds the key to the precise design of rigid-bodied mechanisms~\cite{Jongwon2014Leg}. It is the engineer's choice in design guided by the application needs that lead to the realization of different linkages, where planar linkages and their variations historically play a dominant role in legged, armed, and fingered systems~\cite{Gavin2016DesignPrinciples}. However, when tracing back to the kinematic origins of a planar linkage with four revolute joints (4\textit{R}), which is derived from a loop-closure formulation, one would quickly realize that there are alternative solutions beyond the planar case~\cite{Song20116RLinkage}. These alternatives are not widely utilized in modern robotic limb designs.
    
    The loop closure equation for a 4\textit{R} linkage suggests three possible solutions~\cite{Gu2023OCLimbDesign}, including the overconstrained 4\textit{R} (or the Bennett linkage with skewed joint axes and non-zero-length links), a planar 4\textit{R} (with all parallel joint axes and non-zero length links), and a spherical 4\textit{R} (with all intersecting joint axes and zero-length links). The overconstrained case could be considered a more generalized solution concerning the parametric specification order in geometric conditions. By specifying all twist angles or link lengths as zeros, the Bennett linkage effectively degenerates into a planar or spherical one, respectively. However, it should be noted that one can further generalize the corresponding planar or spherical design by relaxing the equal link lengths or twist angles in the opposite link-pairs, formulating extended design variations of 4\textit{R} linkages in planar and spherical cases. 

    While distributed actuation directly attached to each joint in a kinematic open-chain is straightforward in engineering~\cite{Hutter2016ANYMAL}, limitations in power density and cost constraints usually prefer the adoption of linkages for power and motion transmission~\cite{Zhong2019LegReview}, where the actuators could be placed closer to the robot's main body for a much-reduced inertia~\cite{Gao2020Me}. The design of robotic limbs, including robotic arms, legs, and fingers, involves extensive examples utilizing such design principles. For example, the Ghost Robotics Minitaur~\cite{Tan2018Sim2Real} and Stanford Doggo~\cite{Kau2019Doggo} adopt co-axially arranged dual actuators on the robot's body frame to power a planar 4\textit{R} linkage in a diamond shape as robotic legs. Recently, a more compact design has been widely adopted by adding another motor orthogonally arranged to two co-axial motors to provide spatial motions, where the leg designs are optimized within a tight form factor using a very slender planar 4\textit{R} linkage for power transmission~\cite{Kalouche2017Goat}, or using belt~\cite{Bledt2018Cheetah} or chain~\cite{Hutter2013Legs} transmission to achieve variational designs of planar 4\textit{R} transmission similarly. Hydraulic actuation is also a possible design, which is kinematically equivalent to a variational design of planar 4\textit{R} transmission~\cite{zong2024bionic, RAIBERT200810822}. The parallel mechanism is also a possible design with much increased load-bearing capability but requires significantly more components in assembly~\cite{Gao2020Me, Kalouche2017Goat, Hubicki2016ATRIASDA}. Emerging designs also leverage material softness to develop novel soft lower limbs but generally suffer from dynamic agility in action~\cite{yao2024multimodal}.

    Extensively studied among theoretical kinematicians, mechanism designers, and robotics researchers, overconstrained linkages exhibit rich and complex geometrical foundations in developing advanced mathematical tools for mechanical and robotic systems yet under-adopted in modern applications~\cite{waldron1979overconstrained, GAN20133rTPSMMetamorphicParallelMechanism, Kang2017PneumaticMuscle, Ghafoor2004StiffnessModeling, Dai1995FiniteTwistMapping, LIU2003Carton-folding}. Recent research shows that the overconstrained linkages could also be adopted for developing novel robotic limbs with reconfigurable capabilities, where the overconstrained geometry enables a workspace in 3D surfaces that could be applied for legged locomotion actuated by only two motors per limb~\cite{Feng2021OverconstrainedLeg}. The theoretical foundation and engineering implementation of overconstrained linkages in 4\textit{R}, 5\textit{R}, and 6\textit{R} formulations have been systematically established for the application as robotic limbs in legged locomotion with proven superiority in omnidirectional walking, especially when walking sideways or turning-on-the-spot~\cite{Gu2022Overconstrained}. Recently, the researchers proposed a computational optimization framework to co-design the overconstrained robotic limb while considering the efficiency in quadrupedal locomotion, where the overconstrained limbs are found to be the most efficient in mechanical energy in general~\cite{Gu2023OCLimbDesign}. As a result, a novel class of overconstrained locomotion is emerging, which originates from the incorporation of overconstrained linkages as the robotic limbs for power and motion transmission, which is found to be energy efficient using model-based methods when walking on flat terrains in simple tasks.

    Advanced control methods are required for dynamic walking on challenging terrains, where \textit{model-based} and \textit{learning-based} methods are widely adopted~\cite{qi2021perceptive, ICRA2020b, castillo2021robust, ICRA19b}. Model-based methods use detailed, pre-set mathematical models to understand the robot's movement and surroundings~\cite{qi2021perceptive, castillo2021robust}. They work well in austere, predictable environments but can struggle in more complicated or changing landscapes~\cite{eth2022scirobotics}. On the other hand, learning-based methods, especially reinforcement learning, give a policy guiding robots to adjust and learn from their experiences~\cite{eth2022scirobotics, xu2022polytopic}. This added more flexibility, as pre-defined models do not limit them and can adapt to various and variant terrains and obstacles~\cite{eth2022scirobotics}. For example, using reinforcement learning, cross-skill generalization of overconstrained robotic limbs between manipulation and locomotion was recently found to be an effective method with a co-training framework~\cite{Sun2023LocoManipulation}. Recently, a large-scale, multi-terrain reinforcement learning method has been proposed and deployed successfully in robotic hardware over challenging terrains in the wild~\cite{Rudin2022Learning}. 

    In this paper, we propose a design and learning method to verify the capabilities of overconstrained locomotion in challenging terrains using large-scale, multi-terrain reinforcement learning against the parametric reconfiguration of evolutionary morphology. The resultant design of the overconstrained robotic limb can be parametrically reconfigured to mimic the mammal-inspired and reptile-inspired foothold during walking, representing two major classes of tetrapod limb morphology. Comparative analysis has been found difficult in previous work due to the different robot designs, which becomes possible in this study using limbs with parametrically reconfigurable designs attached to the same robot body. Instead of previous work tested using model-based methods in simple walking tasks on the floor, this research delves further into dynamic walking over a series of challenging terrains, including flat floors, slopes, and stairs, with or without noises in upward or downward directions, all trained via data-driven method using reinforcement learning. The main contributions of this paper are listed as the following: 
    \begin{itemize}
        \item Achieved dynamic walking over a series of challenging terrains by training overconstrained locomotion via large-scale, multi-terrain deep reinforcement learning;
        \item Discovered the superior energy efficiency of overconstrained locomotion over the widely adopted planar limbs and novel spherical limbs in dynamic quadrupedal locomotion via a learning-based approach. 
        \item Further trained a wheel-legged overconstrained locomotion to verify the versatile capabilities of overconstrained robotic limbs in modern legged robots. 
    \end{itemize}

    The rest of this paper is organized as follows. Section~\ref{sec:Method} presents our additive design of overconstrained robotic limbs, the evolutionary morphology of overconstrained quadruped, and our proposed reinforcement learning framework and the simulation setup. Section~\ref{sec:Results} depicts simulation results. Discussions are included in section~\ref{sec:Discuss}, with limitations and future works in the final section.

%%%%%%%%%%%%%%%%%%%%%%%%%%%%%%%%%%%%%%%%%%%%%%%%%%%%%%%%%%%%%%%%%%%%%%%%%%%%%%%%
\section{Method}
\label{sec:Method}
%%%%%%%%%%%%%%%%%%%%%%%%%%%%%%%%%%%%%%%%%%%%%%%%%%%%%%%%%%%%%%%%%%%%%%%%%%%%%%%%

    This section introduces an additive design of the overconstrained robotic limbs proposed in this study, which can be conveniently reconfigured parametrically into planar and spherical linkages for comparative analysis, which shares an evolutionary morphology transforming between reptile- and mammal-inspired legged robots. Then, we present the deep reinforcement learning framework adopted for training overconstrained locomotion in large-scale, multi-terrain scenarios with massively parallel deployment of overconstrained robots in simulations.

\subsection{Additive Design of Overconstrained Robotic Limbs}
    
\begin{figure}[ht]
        \centering
        \includegraphics[width=0.6\textwidth]{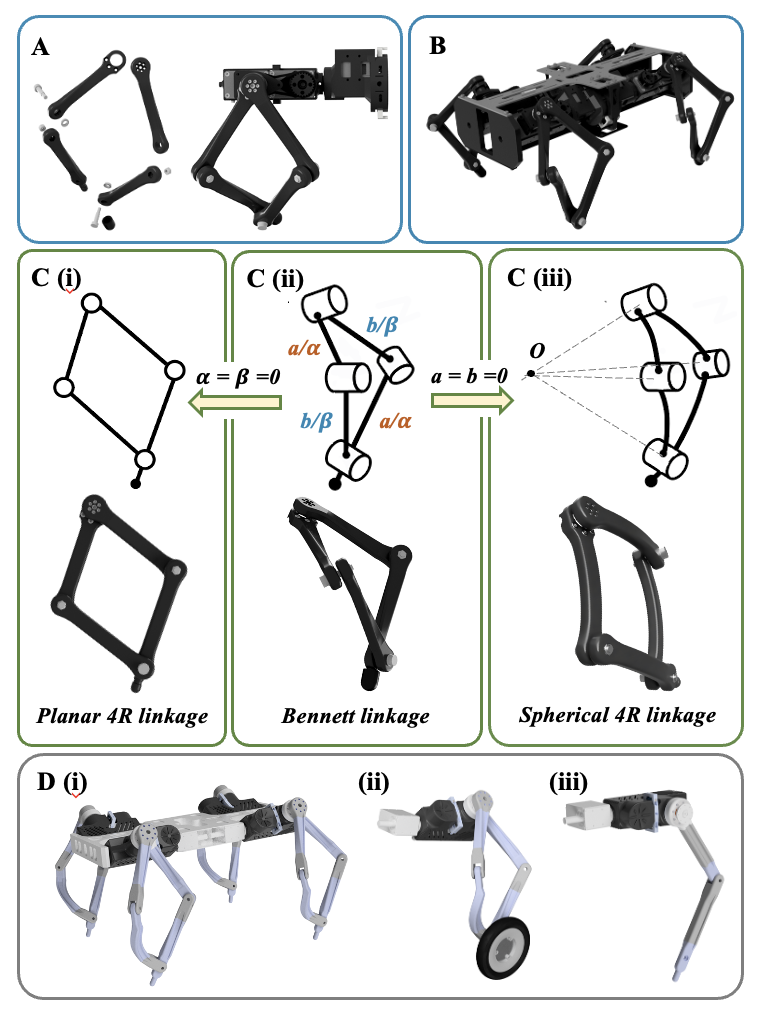}
        \caption{
        \textbf{Parametrically reconfigurable additive design of overconstrained robotic limb.}
        (A) 3D printable components and assembly of an overconstrained robotic limb; 
        (B) Physical prototype of the fully-assembled quadruped in Bennett limbs; 
        (C) Parametric reconfiguration as (i) a planar four-bar, (ii) an overconstrained 4-bar, and (iii) a spherical four-bar; 
        (D) An enhanced design for amphibious locomotion reconfigurable as (i) a legged robot, (ii) a wheel-legged robot, and (iii) a simplified equivalent as an open-chain limb identical if using a belt or slender linkages.
        }
        \label{fig:AddDesign}
    \end{figure}

    Show in Fig.~\ref{fig:AddDesign}A is the overconstrained robotic limb designed for additive manufacturing. All links are designed as the alternative form following the geometric conditions defined by the Bennett linkage. A snap-clutch connector is attached to the first servo motor (DYNAMIXEL Model XM430-W270) with its axis aligned along the body frame, enabling a detachable design of the whole limb to the main body. Two servo motors of the same model are arranged next to each other, where a belt transmission is used to enable a co-axial output connecting two driving links of the Bennett linkage, respectively. Each link is designed by sweeping along the link axis offset by the link's thickness to enable a scissor design for minimum collision during motion. The links will contact each other when they are close, providing a physical limit to its movable range. A ball bearing is fitted inside the joint so that a shoulder bolt can be inserted as the shaft. Each link is 3D-printed using PA-12 (or Nylon 12). Note that to provide enhanced friction to the ground, one of the bottom links is added with an extra extrusion capped with a small piece of rubber at the tip of the toe. When four limbs of the same design are snap-fitted to the body frame, the robot can be assembled quickly, as shown in Fig.~\ref{fig:AddDesign}B. 

    Shown in Fig.~\ref{fig:AddDesign}C(i) is a planar case of the 3D printable design, which can be parametrically modified by changing all twist angles to zeros from a Bennett linkage design shown in Fig.~\ref{fig:AddDesign}C(ii), and then sent for 3D printing and assembly following the same procedure explained above. Alternatively, one can modify the link design so that all joint axes intersect, as shown in Fig.~\ref{fig:AddDesign}C(iii), resulting in a novel robotic limb of spherical linkage. The key advantages of the proposed design are the possibility of rapid reconfiguration in design parameters within the same design for additive manufacturing in 3D printing and direct testing without building a new robot. 

    Shown in Fig.~\ref{fig:AddDesign}D(i) is an enhanced version of the overconstrained quadruped built. Following the same principle, the motors in this design are upgraded to water-proof servo motors (DYNAMIXEL XW540-T140) to provide large torques with amphibious capabilities between on-land and underwater walking. The more robust motors used in this design also enable us to attach a driving wheel to the tip joint of each limb, resulting in a wheel-legged design as shown in Fig.~\ref{fig:AddDesign}D(ii). If two of the linkages are removed, as shown in Fig.~\ref{fig:AddDesign}D(iii), the limb becomes equivalent to the widely adopted design in a pseudo-open-chain, which can be realized using belt or chain transmission for a more compact form factor in design. 
    
    It should be noted that while such differences may seem trivial in design, the simulation of a closed-loop linkage remains challenging in many modern simulation environments or physics engines. For example, a classical problem with the widely adopted robot modeling language URDF (Universal Robot Description Format) is the lack of support for simulating closed-loop linkages. The historical discussion of this challenge is out of the scope of this paper. We have custom-developed and open-sourced a toolbox for this study to address this issue~\cite{qiu2023describing}, enabling seamless conversion of closed-loop linkage designs in CAD software with modern learning-based simulation environments such as PyBullet, DeepMind's MuJoCo, and NVidia's Issac Sim. However, it should be noted that the adaptation for NVidia's Isaac Gym remains an unresolved issue.
    
\subsection{Evolutionary Morphology of Overconstrained Quadruped}

    The resultant quadrupeds achieved based on the reconfigurable overconstrained robotic limbs shown in Fig.~\ref{fig:AddDesign}C provides an interesting perspective into the morphological evolution of tetrapods, especially between mammals and reptiles. During the biological evolution of the limbs, the reptiles were among the first evolved morphology of the tetrapods~\cite{Clack2009FintoLimb}, where the footholds, or the width of the spread between a pair of lower limbs, are usually much wider than the body width. While mammalian evolution has many characteristics, the mammal's footholds are usually comparable to its shoulder width. 

    Such morphonological evolution could also be observed from the kinematics perspective by reconfiguring the 4\textit{R} linkages as the robotic limbs. For example, the overconstrained 4\textit{R} limb shown in Fig.~\ref{fig:AddDesign}C(ii) would result in quadruped locomotion where the footholds are much more comprehensive than the body width, which are similar to the reptiles during walking. Classical examples such as frogs, ants, spiders, and lizards exhibit similar morphology during walking. However, when fixing all twist angles to zeros, the resultant planar limbs shown in Fig.~\ref{fig:AddDesign}C(i) exhibit a closer resemblance to the mammal, where the footholds are expected to be comparable to the body width. For example, the width of the spread of the two human feet, or the foothold, is roughly comparable to the width of the shoulder. One would be diagnosed as physically impaired if the foothold is too narrow, resulting in in-toeing behaviors, or when the foothold is too broad, resulting in splayfoot symptoms. This is also widely observed in other mammals, such as cheetahs, elephants, etc.

    Mammals such as cheetahs are usually faster but cannot sustain long-distance locomotion like the reptiles such as lizards. When building quadrupedal robots, a balancing choice of design is critically important to develop legged systems that are more agile in speed or more energy-efficient over challenging terrains. Addressing this research question would require building a reconfigurable quadrupedal system that can be conveniently switched between different footholds based on a comparable kinematic architecture, making the proposed system shown in Fig.~\ref{fig:AddDesign} an effective solution for further investigation. Furthermore, instead of pursuing a model-based control, which may result in hand-engineered differences in parameter selection, a learning-based approach shows promising effectiveness in finding optimal locomotion despite the morphological differences via a data-driven approach, which will be addressed next.

\subsection{Reinforcement Learning of Overconstrained Locomotion}

    \subsubsection{Simulation Setup for Legged Locomotion}

\begin{figure}[ht]
        \vspace{2mm}
        \centering
        \includegraphics[width=1\textwidth]{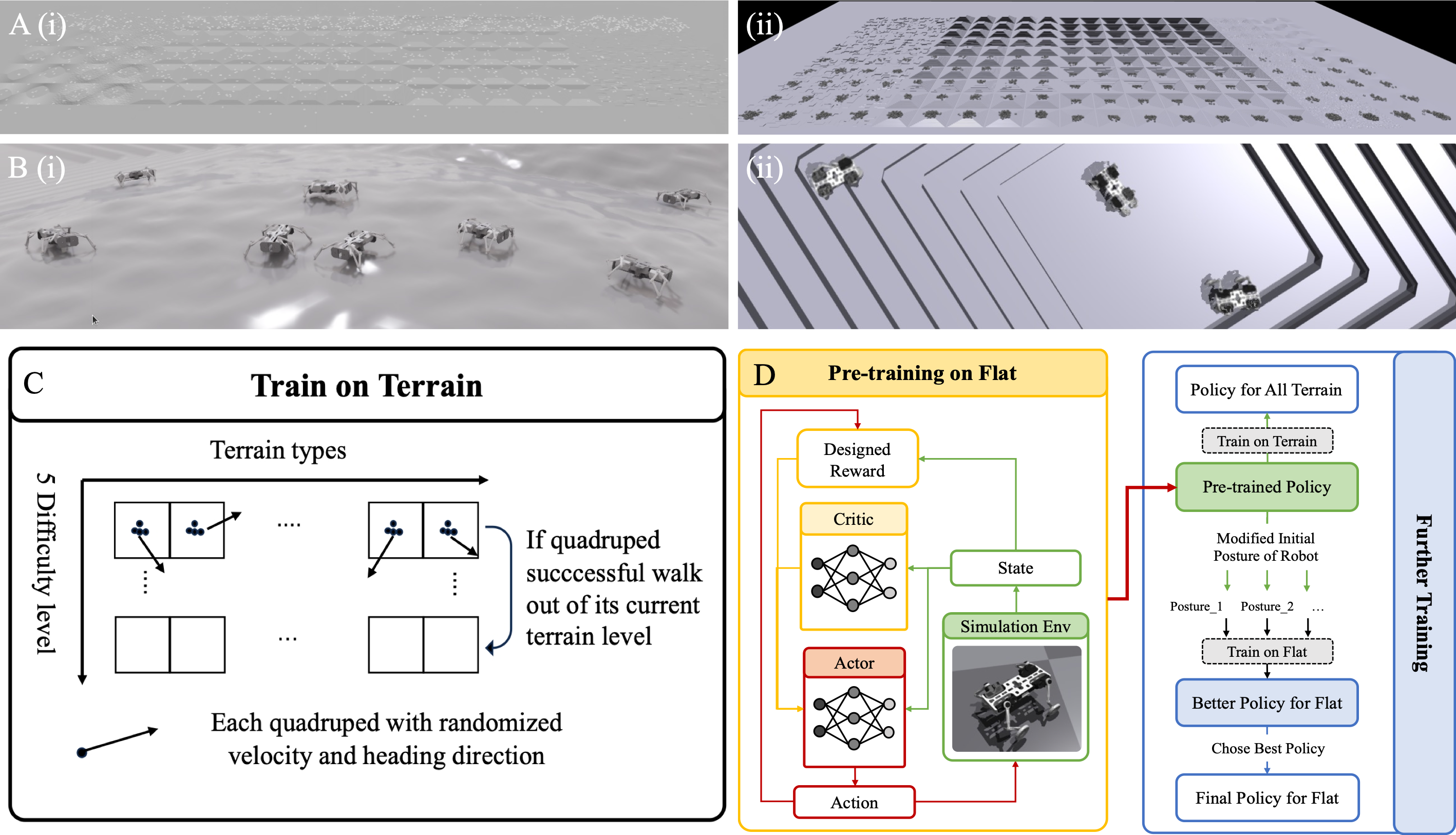}
        \caption{
        \textbf{Simulation setup for massively parallel deep reinforcement learning.}
        (A) Screenshot of the training scene for reinforcement learning in a large-scale, multi-terrain environment using (i) 2048 overconstrained-legged robots and (ii) 4096 overconstrained wheel-legged robots. 
        (B) Closed-up view of the training scene with (i) overconstrained-legged robots and (ii) overconstrained wheel-legged robots;
        (C) Schematic of the multi-terrain training process; 
        (D) Flow chart for training overconstrained locomotion.
        }
        \label{fig:Method-SetupRL}
    \end{figure}

    We adopted the setup by Isaac Sim to train overconstrained locomotion via massively parallel deep reinforcement learning on a single workstation GPU~\cite{Rudin2022Learning}. The goal is to train the robot to navigate rugged terrain with a practical curriculum-based approach that is well-suited for simultaneous simulations with thousands of robots. The simulation environment setup is shown in Fig.~\ref{fig:Method-SetupRL}A, a large scene with 20-by-5 grids surrounded by flat floors extended to the edge of the scene. Each grid is a terrain type of 8-by-8m square. Horizontally, there are 20 terrain blocks of different types, including smooth slope, rough slope, stairs up, stairs down, and discrete, and the proportions of each terrain type are 0.1, 0.1, 0.35, 0.25, 0.2. Vertically, there are terrains of the same type with increasing difficulties when changing the corresponding terrain parameters. The simulation process is like a game, with 2048 robots evenly divided into groups deployed randomly at the center of the most manageable level of terrain blocks. Each robot is given command of random linear velocity and heading direction $[v_x,v_y,\theta_h]$ between -1 and 1 m/s to learn how to walk out of each grid onto the next. We adopted similar reward terms as in~\cite{Rudin2022Learning} but excluded the joint torques penalty to best use the servo motors' relatively limited torque output.

    \subsubsection{Simulation Setup for Wheel-Legged Locomotion}    

    We adopted the same massively parallel deep reinforcement learning to train the wheel-legged locomotion. We prioritize key reward terms for learning locomotion policies. Each term, weighted to emphasize velocity tracking, joint motion minimization, and collision avoidance, is detailed in the provided appendix. We initially trained a pre-trained model on a flat floor in our Legged Gym pipeline. After selecting the optimal pre-trained policy, we further train the model on diverse terrains. If the outcomes fall short of expectations, we adjust the robot's initial posture and return to the flag. In the experiment, we trained a policy for our design with over-constrained wheel legs, sequentially progressing from flat floors to more challenging terrains.

    Since the Isaac Gym does not support closed-loop linkage simulation, we transfer the overconstrained leg into a pseudo-open chain, as shown in Fig.~\ref{fig:AddDesign}D(ii). In planar 4-bar closed chains, symmetry determines the entire chain's state once the left or right two linkages' states are known. When emphasizing the shoulder-to-foot position constraint, other constraints become negligible. This concept can be extended into the spatial 4R linkages, such as the Bennett mechanism. 

%%%%%%%%%%%%%%%%%%%%%%%%%%%%%%%%%%%%%%%%%%%%%%%%%%%%%%%%%%%%%%%%%%%%%%%%%%%%%%%%
\section{Results}
\label{sec:Results}
%%%%%%%%%%%%%%%%%%%%%%%%%%%%%%%%%%%%%%%%%%%%%%%%%%%%%%%%%%%%%%%%%%%%%%%%%%%%%%%%

    This section starts by presenting the reinforcement learning results for overconstrained locomotion in comparison against planar and spherical limbs. Then, we present the simulated testing results measured in various metrics, paying particular attention to the Cost-of-Transport analysis. Finally, we report the reinforcement learning results for overconstrained wheel-legged locomotion.

\subsection{Overconstrained Legged Locomotion over Multi-Terrains}
\label{sec:Result1-OverRL}
    
\begin{figure}[ht]
        \centering
        \includegraphics[width=0.6\textwidth]{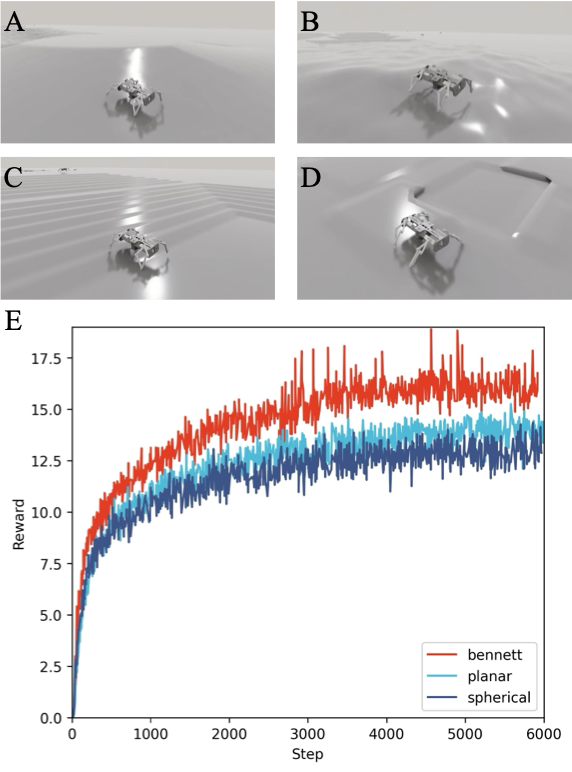}
        \caption{
        \textbf{Large-scale, multi-terrain reinforcement learning of overconstrained locomotion.}
        Besides walking on the simple flat floor, here are the screenshots of the overconstrained quadruped walking on (A) simple slopes, (B) slopes with random noises, (C) stairs, and (D) flat floors with random noises. (E) reports the reinforcement learning rewards of the quadruped with Bennett, planar, and spherical limbs over 6K training steps.
        }
        \label{fig:OverRL}
    \end{figure}

    We begin by conducting large-scale, multi-terrain deep reinforcement learning of overconstrained locomotion using the setup illustrated in Fig.~\ref{fig:Method-SetupRL}. Figs.~\ref{fig:OverRL}A-D are the screenshots of the trained robot in overconstrained locomotion in various terrains. The rewards are shown in Fig.~\ref{fig:OverRL}E, where the quadruped with Bennett limbs in red lines indicate that the overconstrained locomotion is successfully trained within about 3,000 steps and saturated with little increase when trained to 6,000 steps and beyond. Using the identical setup, we also trained the quadruped with planar and spherical limbs, plotted in the light and dark blue lines in Fig.~\ref{fig:OverRL}E, respectively. 

    Interestingly, the Bennett limbs exhibit the best training result in reinforcement learning, indicating the superior performance of overconstrained locomotion over dynamic walking with planar or spherical limbs. While also being a rather unconventional design in legged robots that is yet to be explored in literature, the spherical limbs exhibit only a slight decrease in dynamic walking performance compared to the planar limbs, as indicated in our simulation results. The spherical linkages are widely adopted in surgical robots but have yet to be applied in legged robots, which may provide further benefits by leveraging its remote center of workspace, which is outside the scope of this study. 
    
\subsection{Comparing Limb Morphology in Locomotion Efficiency}
\label{sec:Result2-Energy}

\begin{figure}[ht]
        \centering
        \includegraphics[width=0.65\textwidth]{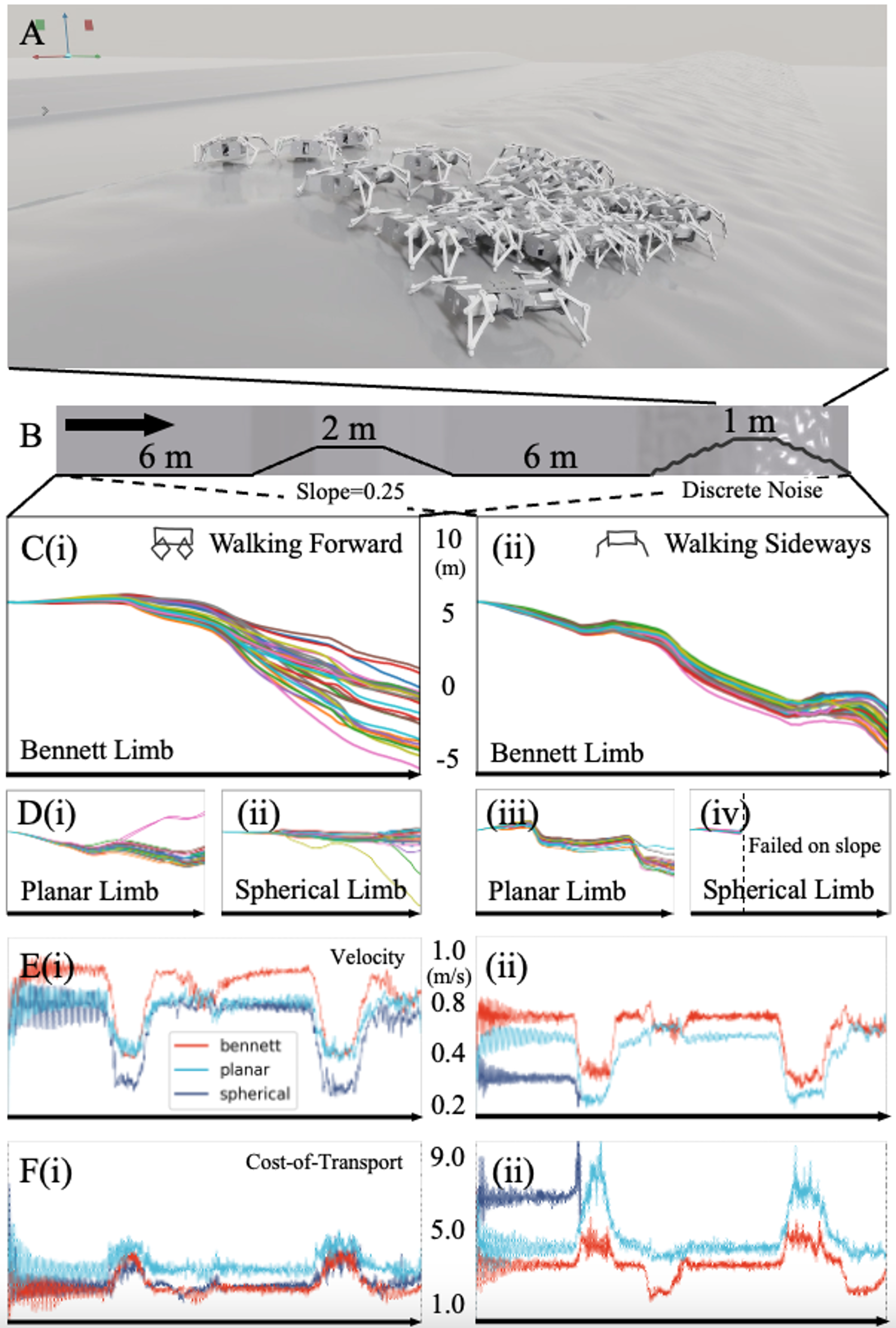}
        \caption{
        \textbf{Performance benchmark on terrain traverse task.}
        (A) 30 robots with Bennett limbs are completing the traverse task by walking sideways; 
        (B) Terrain distribution of the traverse task used to benchmark the locomotion policies; 
        (C) Trajectories of quadrupeds with Bennett limbs walking forward (i) and sideways(ii); 
        (D) Trajectories of quadrupeds with planar or spherical limbs walking forward (i, ii) and sideways(iii, iv);
        (E) Average horizontal linear velocity of quadrupeds concerning the distance walked along the traverse direction;
        (F) Average Cost-of-Transport concerning the distance walked along the traverse direction.
        }
        \label{fig:Energy}
    \end{figure}

    We benchmarked the performance of the three trained locomotion policies for Bennett, planar, and spherical limbs. The goal of the task is to let the quadruped traverse across a series of two flat, one slope, and one rough slope terrains, as shown in Fig.~\ref{fig:Energy}A.

    Each terrain block is 6 meters in distance, uniform along the \textit{x}-axis, and varies along the \textit{y}-axis. The sloping terrain has a rate of 0.25 and a plateau of 2 meters. The rough terrain increases the task difficulty by adding a randomly uneven surface and a higher plateau at 0.625 meters. For each limb design, thirty quadrupeds were initiated with the same pose at the starting point and commanded to walk forward or sideways toward the \textit{y}-axis at the speed of 1 m/s. The actual forward and sideways trajectories are shown in Fig.~\ref{fig:Energy}C and D. All robots with Bennett or planar limbs completed the traverse task by walking forward and sideways. However, all robots with spherical limbs completed the traverse task by walking forward but failed to climb the sloping terrain.

    Fig.~\ref{fig:Energy}E shows the horizontal linear velocity averaged over 30 quadrupeds over different terrains. Quadruped with Bennett limbs is the fastest on all test terrains, especially on the flat terrain where Bennett one achieves an average forward speed of 0.85 m/s, making a 20\% increase compared to the planar one's at 0.71 m/s. The average time to complete the task by walking forward is 35.45, 38.28, and 44.91 s for Bennett, planar, and spherical limbs. We also investigate the Cost-of-Transport (COT) defined by $P/mgv$ for each limb design. The Bennett limb has the lowest COT over all tested terrains in forward and sideways walking, indicating the highest energy efficiency. The accumulated mechanical energy used to complete the task is reduced by 22.4\% and 30.6\% compared to the planar limb in forward and sideways walking, respectively.

\subsection{Learning Wheel-Legged Overconstrained Locomotion}
\label{sec:Result3-WheelLeg}
    
\begin{figure}[ht]
        \centering
        \includegraphics[width=0.9\textwidth]{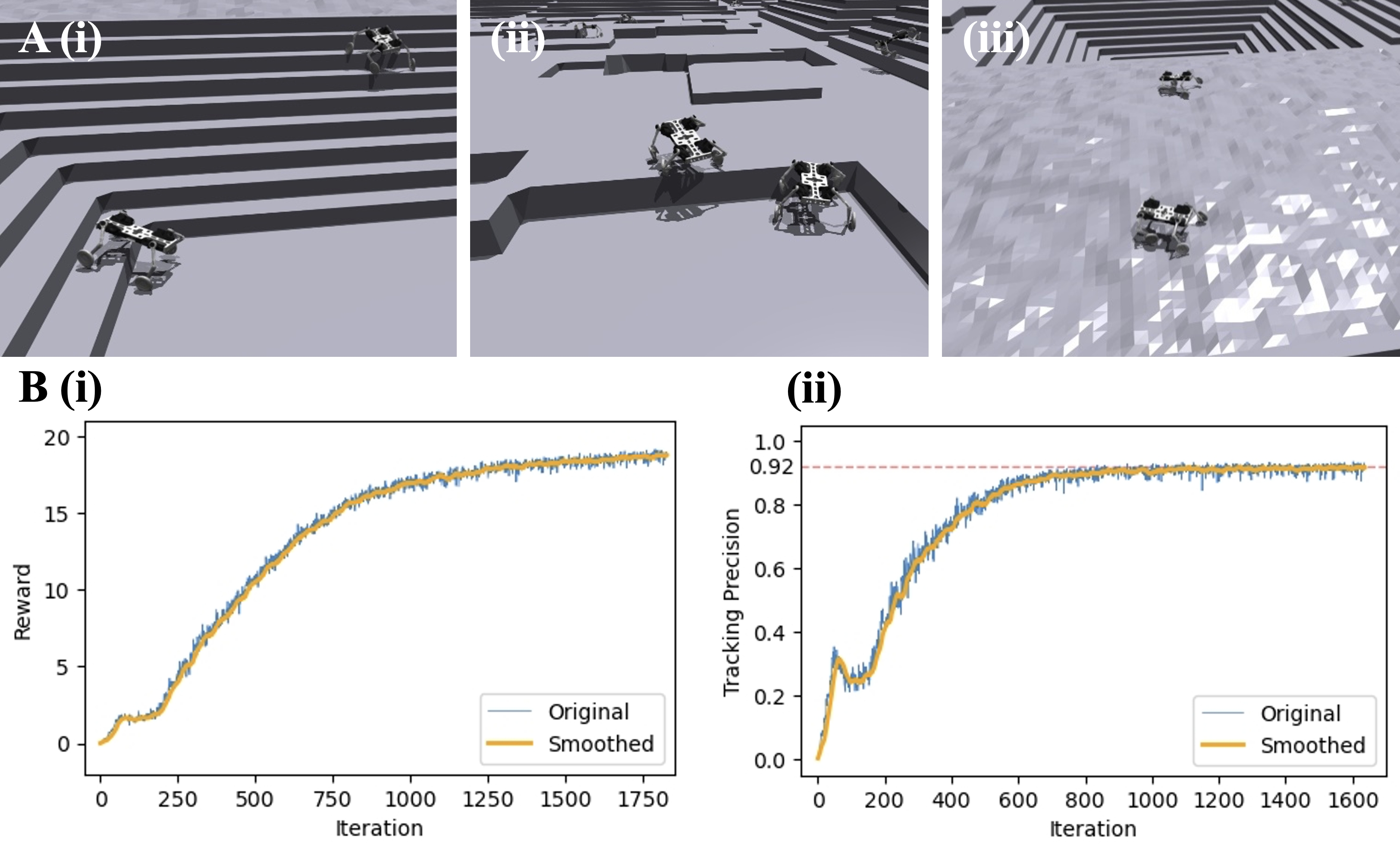}
        \caption{
        \textbf{Learning wheel-legged overconstrained locomotion.}
            (A) Testing Policy on Complex Terrain:
                (i) Evaluating performance on stairs.
                (ii) Assessing obstacle navigation capabilities through random obstacles.
                (iii) Evaluating adaptability on uneven and rugged terrain.
            (B) Training Data Analysis Throughout Training Process:
                (i) Mean reward progression over time steps.
                (ii) The precision of tracking linear velocity commands.
        }
        \label{fig:Result3-WheelLeg}
    \end{figure}

    Quadrupeds equipped with the overconstrained Bennett wheel legs undergo comprehensive training across diverse terrains, focusing on assessing their performance on flat floors. The findings are presented in Fig.~\ref{fig:Result3-WheelLeg}, providing a comprehensive understanding of the quadrupeds' locomotive capabilities. We have verified that reinforcement learning can be utilized to train a control policy for locomotion tasks. We test commands on forward, backward, left, and right movement, and the robot can follow up well in flat terrains. Also, we test the same commands on various other terrains. The locomotion performance remains good on the discrete flat, rough slope, slope terrains, and going downstairs. However, it currently cannot go upstairs under such a policy.

%%%%%%%%%%%%%%%%%%%%%%%%%%%%%%%%%%%%%%%%%%%%%%%%%%%%%%%%%%%%%%%%%%%%%%%%%%%%%%%%
\section{Discussions and Conclusion}
\label{sec:Discuss}
%%%%%%%%%%%%%%%%%%%%%%%%%%%%%%%%%%%%%%%%%%%%%%%%%%%%%%%%%%%%%%%%%%%%%%%%%%%%%%%%

\subsection{Learning Superior COT in Overconstrained Locomotion}

    This study presents the first learning-based investigation of overconstrained locomotion over challenging terrains, providing data-driven evidence that supports the superior energy efficiency of overconstrained robotic limbs over the widely adopted planar limbs and another novel class of spherical limbs. Previous research adopted a model-based approach to computationally co-design the overconstrained robotic limb for the most mechanical efficiency in omnidirectional walking on the floor. This study successfully implemented dynamic walking with overconstrained robotic limbs over a wide range of challenging terrains, including floors, slopes, and stairs, with or without random noises, where the overconstrained limbs exhibit consistent superior performance by saving at least 22\% mechanical energy in both forward and sideways traverse tasks. It also achieves the highest average speed of 0.85m/s on flat terrain, which is 20\% faster than the planar limbs. This work indicates the exciting potential of overconstrained linkages in designing modern robotic limbs in dynamic locomotion, which may also apply to dexterous manipulation, as reported in recent literature~\cite{Sun2023LocoManipulation}. This study paves the path for an exciting direction for future research in overconstrained robotics leveraging evolutionary morphology and reconfigurable mechanism intelligence when combined with state-of-the-art methods in deep reinforcement learning.

\subsection{Data-driven Mechanism Intelligence Supporting Evolutionary Morphology with Overconstrained Robotic Limbs}

    Combining reconfigurable intelligence from mechanism science and learning-based behavior training can provide novel solutions for developing evolutionary-inspired robot design methods and reconfigurable systems. Quadrupedal systems are highly integrated, requiring systematic engineering involving multiple aspects, making it challenging to compare robotic limbs with different power and motion transmission mechanisms. In this study, we adopted a reconfigurable method to design the robotic limbs for additive manufacturing, which can be parametrically re-engineered to achieve all three types of 4\textit{R} linkages in planar, overconstrained and spherical cases within a single robot system. In addition, our results found that the mechanism intelligence demonstrated in the generalized geometry of overconstrained linkages over planar and spherical ones shows a similar pattern observed in the morphological evolution of Fin-to-Limb transformation among the animal limbs. The overconstrained limbs resemble the reptile tetrapods with more energy-efficient walking over challenging terrains. At the same time, the existing literature shows that planar limbs originating from mammal tetrapods are generally suited for agile locomotion. Furthermore, our learning-based results indicate that reptile-inspired locomotion with a broader foothold could also be agile in both forward and sideways locomotion at high speed while maintaining a lower Cost-of-Transport compared to the mammal-inspired planar limbs, as demonstrated in the results shown in Figs.~\ref{fig:Energy}E\&F.

\subsection{Learning Wheel-Legged Overconstrained Locomotion}

    Using a similar reinforcement learning framework, we also achieved a novel overconstrained wheel-legged locomotion in simulation, holding the potential for future applications in multi-modal, reconfigurable locomotion in an amphibious environment. While results in this study already established the energy efficiency in overconstrained locomotion, further enhancement could be achieved by introducing a wheel-legged design to the Bennett limb. For traversing tasks on terrains with continuous roughness, wheel-based designs are more efficient for high-speed movement, which can be combined with the legged system for an integrated design. However, our results also indicate that the wheel-legged system is more challenging to implement in reinforcement learning, requiring further research and system design. The added weight of the extra motors and wheels could pose further challenges to the motor selection of the robotic limb. Nevertheless, our simulation results further establish the possibility of refining overconstrained robotic limbs with the state-of-the-art wheel-legged system design for developing advanced, agile, and energy-efficient legged systems. 
    
%%%%%%%%%%%%%%%%%%%%%%%%%%%%%%%%%%%%%%%%%%%%%%%%%%%%%%%%%%%%%%%%%%%%%%%%%%%%%%%%
\section{Limitations and Future Works}
\label{sec:Final}
%%%%%%%%%%%%%%%%%%%%%%%%%%%%%%%%%%%%%%%%%%%%%%%%%%%%%%%%%%%%%%%%%%%%%%%%%%%%%%%%

    We acknowledge several limitations of our study. Verification through hardware implementation must be conducted to consolidate the results reported in this study, as this study is mainly conducted through simulation. The Sim-to-Real gap could pose a significant challenge to effectively deploying the learned skills presented in this study. The current development of learning-based simulation environments is still mainly suited for robotic limbs with open chains, which requires further development so that closed-loop mechanisms can be more effectively and efficiently simulated in large-scale, multi-terrain reinforcement learning. Considering that the prototype is mainly driven by servo motors with limited torque density and control bandwidth, actual performance in energy efficiency requires further experimentation and testing. Also, in the wheel-legged simulation, even with a more powerful servo motor attached, we had to increase the torque output by 50\% to train the overconstrained wheel-legged locomotion. This suggests that a systematic re-selection of the motors might be necessary to build such a dynamic system to ensure sufficient torque reserve for dynamic walking and terrain traverse tasks. We have already built another full-sized prototype of the overconstrained quadruped using water-proofed quasi-direct drives with much-enhanced torque output, which we plan to develop further to test the results obtained in this study. A trade-off of this study by adopting a reconfigurable limb design suggests that there is still room for further enhancement in optimizing the overconstrained robotic limb if used for a singular purpose without considering reconfiguration into other linkage topologies. Future research of this study will focus on developing an overconstrained quadruped for cross-limb, cross-terrain, and cross-skill in an amphibious environment with autonomous morphological reconfiguration. 

%%%%%%%%%%%%%%%%%%%%%%%%%%%%%%%%%%%%%%%%%
\section*{Acknowledgment}
%%%%%%%%%%%%%%%%%%%%%%%%%%%%%%%%%%%%%%%%%

    This work was partly supported by the National Natural Science Foundation of China [62206119, 52335003], the Science, Technology, and Innovation Commission of Shenzhen Municipality [JCYJ20220818100417038], Shenzhen Long-Term Support for Higher Education at SUSTech [20231115141649002], SUSTech Virtual Teaching Lab for Machine Intelligence Design and Learning [Y01331838], and the SUSTech Robotics Club.

%%%%%%%%%%%%%%%%%%%%%%%%%%%%%%%%%%%%%%%%%%%%%%%%%%%%%%%%%%%%%%%%%%%%%%%%%%%%%%%%
\newpage
\section*{APPENDIX}
\label{sec:Appendix}
%%%%%%%%%%%%%%%%%%%%%%%%%%%%%%%%%%%%%%%%%%%%%%%%%%%%%%%%%%%%%%%%%%%%%%%%%%%%%%%%
\subsection*{Reward Terms and Descriptions}

    \begin{tabular}{p{2.5cm}|p{13cm}}
        \hline
        \textbf{Reward Term} & \textbf{Description} \\
        \hline
        soft\_dof\_pos\_limit & Penalize degrees of freedom (DOF) positions too close to their limits. \\
        \hline
        soft\_dof\_vel\_limit & Penalize degrees of freedom velocities too close to their limits. \\
        \hline
        base\_height\_target & Reward for maintaining the base height close to the target value. \\
        \hline
        only\_positive\\ \_rewards & Clips negative total rewards at zero to avoid early termination problems. \\
        \hline
        tracking\_sigma & Controls the spread of the tracking reward, where the reward is computed based on the exponential of the error squared divided by the tracking sigma. \\
        \hline
        soft\_torque\_limit & Penalize excessive torques exerted by the robot. \\
        \hline
        max\_contact\_force & Penalize contact forces above this value. \\
        \hline
        termination & Terminal reward or penalty. \\
        \hline
        tracking\_lin\_vel & Reward for tracking linear velocity commands in the x \& y axes. \\
        \hline
        tracking\_ang\_vel & Reward for tracking angular velocity commands in the yaw axis. \\
        \hline
        lin\_vel\_z & Penalize excessive base linear velocity in the z-axis. \\
        \hline
        ang\_vel\_xy & Penalize excessive base angular velocity in the x \& y axes. \\
        \hline
        orientation & Penalize deviations from a flat base orientation. \\
        \hline
        torques & Penalize excessive joint torques exerted by the robot. \\
        \hline
        dof\_vel & Penalize degrees of freedom velocities. \\
        \hline
        dof\_acc & Penalize degrees of freedom accelerations. \\
        \hline
        base\_height & Penalize deviations of the base height from the target. \\
        \hline
        feet\_air\_time & Reward for long steps penalizes excessive time spent with feet off the ground. \\
        \hline
        collision & Penalize collisions on selected bodies. \\
        \hline
        feet\_stumble & Penalize instances where feet hit vertical surfaces. \\
        \hline
        action\_rate & Penalize changes in actions. \\
        \hline
        stand\_still & Penalize motion at zero commands. \\
        \hline
    \end{tabular}
%%%%%%%%%%%%%%%%%%%%%%%%%%%%%%%%%%%%%%%%%
\bibliographystyle{unsrt}
\bibliography{References}  %%% Remove comment to use the external .bib file (using bibtex).
%%%%%%%%%%%%%%%%%%%%%%%%%%%%%%%%%
\end{document}